# Discovery of structure-property relations for molecules via hypothesis-driven active learning over the chemical space

Ayana Ghosh,[1] Sergei V. Kalinin[2] and Maxim A. Ziatdinov[1,3]

[1] Computational Sciences and Engineering Division, Oak Ridge National Laboratory, Oak Ridge, TN 37831 USA

[2] Department of Materials Science and Engineering, University of Knoxville, Knoxville, TN 37996 USA

[3] Center for Nanophase Materials Sciences, Oak Ridge National Laboratory, Oak Ridge, TN 37831 USA

Discovery of the molecular candidates for applications in drug targets, biomolecular systems, catalysts, photovoltaics, organic electronics, and batteries, necessitates development of machine learning algorithms capable of rapid exploration of the chemical spaces targeting the desired functionalities. Here we introduce a novel approach for the active learning over the chemical spaces based on hypothesis learning. We construct the hypotheses on the possible relationships between structures and functionalities of interest based on a small subset of data and introduce them as (probabilistic) mean functions for the Gaussian process. This approach combines the elements from the symbolic regression methods such as SISSO and active learning into a single framework. The primary focus of constructing this framework is to approximate physical laws in an active learning regime toward a more robust predictive performance, as traditional evaluation on hold-out sets in machine learning doesn't account for out-of-distribution effects and may lead to a complete failure on unseen chemical space. Here, we demonstrate it for the QM9 dataset, but it can be applied more broadly to datasets from both domains of molecular and solid-state materials sciences.

email: ghosha@ornl.gov

**Introduction**

Chemical discovery[1-4] is rooted in quantitative structure-activity/property relationships (QSAR/QSPR).[5-12] These efforts primarily rely on finding appropriate representation of molecules followed by establishing relationships between structure and activity they exhibit. These models are harnessed to explore chemical space to select molecules of interest [13-15] for drug targets,[16-20] antibiotics,[21] catalysts,[22-23] photovoltaics,[24-27] organic electronics,[28] redox-flow batteries.[29] In addition, chemical discovery also includes understanding of chemical processes such as reaction energy pathways,[30-32] optimization of reaction conditions,[33] (for e.g., catalytic activity[34]), crystallization,[35-36] docking,[37] and synthesis.[38-39]

The QSAR/QSPR techniques have proven to be useful in all (not limited to) such scenarios. The popularity[40-42] remains in their simplicity to incorporate structural information combined with physicochemical properties, reliability to capture the property landscape, capability to identify existing chemical patterns, and identify activity cliffs within the data while being computationally affordable to perform. The descriptors or features can be multi-dimensional descriptors capturing electronic or topological characteristics. Alternatively, these can be fingerprints that are the effective representations of molecules via graph-based or string representations (SMILES,[43] SELFIES[44]). QSAR/QSPR models began its journey almost 60 years ago with the seminal work lead by Hansch et al.[45] in which a few simple descriptors were used to capture a 2D structure-activity relationship. Since then, this field has seen a steep rise in utilization of variety of traditional ML algorithms (Naïve Bayes, Support Vector Machine, Random Forest, to name a few) for property/process prediction followed by validation by experimental synthesis. Its success is also credited to generation of easily accessible public repositories (e.g., PubChem,[46-48] ZINC,[49] ChEMBL,[50-51] QM9,[52] ANI-1x,[53] and QM7-X[54]) containing structural and physiochemical properties (computed with quantum mechanical calculations or observed with experiments) on thousands to millions of molecules. Altogether these have paved the path forward in chemical design, discovery with day-to-day applications.

If QSAR/QSPR studies have created a revolution in *in silico* design efforts, applications of deep neural network (NN) algorithms[55-56] have further accelerated this progress. They have enabled efficient usage of the big data for not only finding molecules of interest but also quantify[57-60] molecular interactions, chemical bonding, inverse design of molecules for targets and gain novel insights into mechanisms. However, the performance of any of these models is highly dependent on the quality, quantity, modelability of the datasets. Furthermore, a discovery process necessitates extrapolation of learned correlative relationships onto the previously unseen regions of chemical space.

Correspondingly, the task of generation[61] of new molecules by going beyond the standard (manual) design rules or solutions has gained much attention. Several studies[62] have been reported where different NN-based algorithms are explored to accomplish this challenge. Autoencoders are one of the common models that are being used to encode molecules[63] via complex latent representations to optimize for specific properties and map them back to molecular structures through decoding. Another set of examples include applications of recurrent NN[64] algorithms where molecule generation is treated as a sequencing task and the algorithm is permitted to generate samples at each stage, as informed by the model. There are also studies on using self-attention driven transformer models[65-66] for targeted structure generation. In addition, reinforcement learning strategies[67-68] have been implemented in this context where molecules with multiple target objectives can be found. The modern molecular generative models have transformed standard string representations of molecules towards embedded spaces[69] with information on the entire molecular scaffold.

However, the learning approach behind most of the generative models traverse through latent embeddings. The latter are generally not smooth, precluding direct gradient-based optimization methods. Once trained over full libraries of molecules, the fraction of the space occupied with molecules with useful functionalities is typically small, making their discovery complex. The chemical space is non-differentiable, precluding the gradient-based descent or simple Gaussian processes (GPs)[70-71] based methodologies. ML methods including variational autoencoders aim to construct a suitable low-dimensional and ideally differentiable latent embeddings for the chemical space, allowing for the Bayesian optimization (BO)[72-74] type processes. However, these methodologies to date have been based on either construction of the embedding space for the full library of candidate molecules or finding similar kernel of representation for these molecules for target explorations.

Historically, Gaussian process (GPs) has been used within the active learning and BO, making these processes purely data-driven and non-parametric in nature. It interpolates functional behavior over relatively low-dimensional parameters space. In a classical GP, a kernel function (such as radial basis function kernel) is utilized to define the degree of correlation across the parameters space. The kernel parameters are inferred based on the available data, obtained during exploration-exploitation steps. It does not incorporate any prior information of physical or chemical behavior of the system in the process. It learns the physics of the system from the data itself via kernel function with possibilities of leading to suboptimal results. Consequently, the number of optimization steps necessary to reconstruct functional behavior, even scanning over low parameters space becomes large. More advanced physics-informed kernel functions[75-77] may help in such cases which is an area of active research. However, proper application of this technique in the domain of chemical or physical sciences demands going above the usage of conventional GP in BO framework to model functionalities of interest.[78-81] A series of molecular kernels[82] such as fingerprint kernels (e.g., scalar product kernel, Tanimoto kernel), string kernels (e.g., kernels based on SMILEs, SELFIEs) and graph kernels (typically uses molecular fragments) can be used in GP framework. Once the best-performing kernel is chosen, BO is performed for applications in real world scenarios such as optimization of chemical reactions.

In a recent work, Ziatdinov et al.[83-85] introduced a physics-augmented algorithm for active learning and Bayesian optimization. It combines the flexibility of GP models with physical priors allowing for the hypothesis-driven discovery in ML. To date, it has been applied[86] to the experiments in scanning probe microscopy, providing new insights into the concentration-induced phase transition and identifying domain growth laws in ferroelectric materials.

Here, we extend the concept of hypothesis learning to molecular discovery. We combine it with the compressed-sensing methodologies for identifying relevant structural descriptors and evaluate multiple automatically generated hypotheses with a reward-driven acquisition (similar to the reinforcement learning) to select next evaluation points. Specifically, the hypotheses are selected using the compressed sensing performed on combinations of nonlinear functionalized features to find a list of the most relevant combinations. This step is followed by balancing dimensions (i.e., respecting physics constraints) with respect to the target property to formulate them into feasible equations. Here, we only consider a handful of easily computable features related to property of interest, to keep the hypotheses in a simple form that is easy to calculate. Finally, we evaluate the hypotheses over a wide parameters space to predict functionalities of interest within the active learning loop. We have utilized the QM9 dataset on isolated molecules as a use case to establish this tail of our work.

We note that the primary focus of constructing such a framework remains on learning or approximating generalized physical laws in an active learning regime that are the driving factors behind the target functionality of interest. As a general practice, ML models are often evaluated by their performances on hold-out test/validation sets. However, this does not take into consideration the out-of-distribution effects of real-world data. Therefore, performance on hold-out validation sets in terms of average errors may not be the best artifact to capture how ML framework may perform to predict results on unseen data. In contrast, if the framework is focused on deriving a suitable approximation of generalized laws of physics based on observational data, it can readily be applied to predict the functional behavior of unseen chemical space.

A schematic of the generalized framework detailing the active cruise between design and discovery using hypothesis learning is shown in Figure 1. The results along with estimated uncertainties show a generalized cost-effective way to approximate structure-property relations, applicable to a wide variety of material systems.

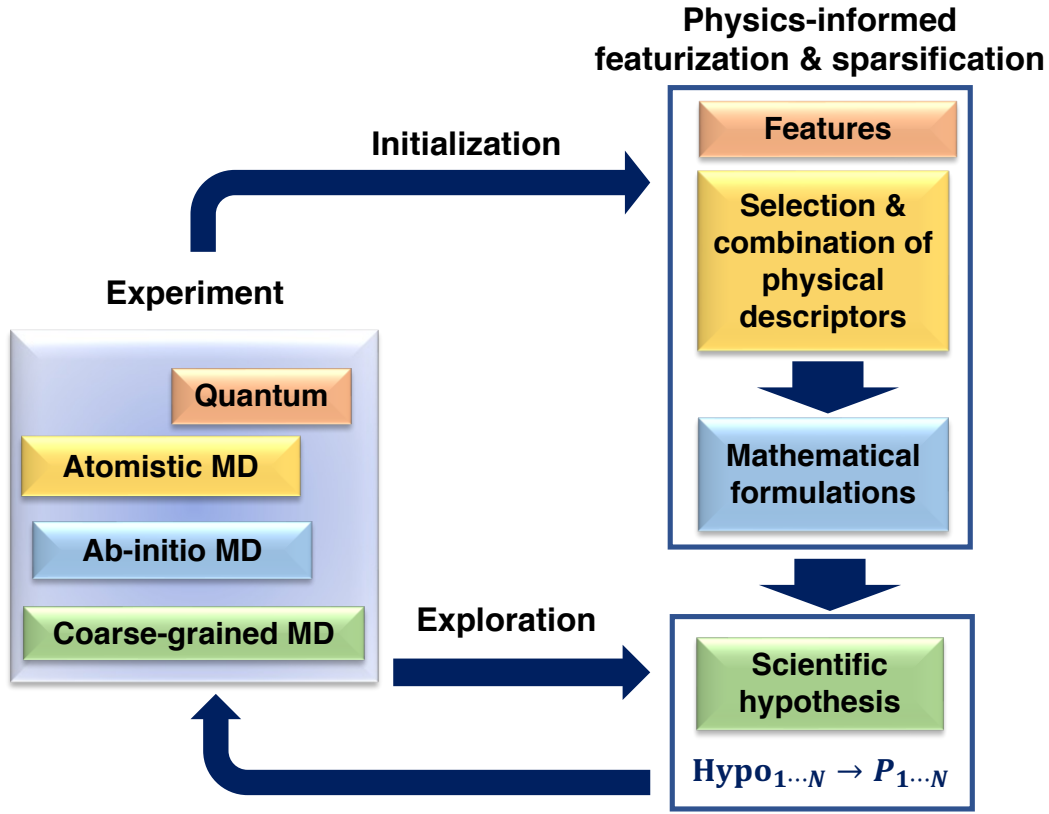

**Figure 1**: **Schematic of workflow, from design to discovery**. Figure (left panel) shows commonly used simulation techniques to generate reliable data for various materials systems. The right panel establishes the active learning loop - combining features to come up with mathematical formulations as statistically derived scientific hypotheses, to be evaluated for discovery structure-property relations.

**Results and Discussion**

**General Considerations:**

The physics-informed featurization scheme that we designed is built upon the compressed sensing methodologies utilized by Ghiringhelli et al.[87] for features selection, implemented here as the seed step for the discovery cycle of an active learning process. Similar to the original SISSO implementation, our physics-informed featurization and sparsification scheme allows for the selection of the most relevant descriptors which is obtained by using the least absolute shrinkage and selection operator (LASSO). The LASSO algorithm employed as a part of feature selection scheme uses the sparsity of the $l_1$ norm to effectively reduce a descriptor set to the most relevant descriptors ($d_i$) contained in full set ($D$). It selects the non-zero terms of the $l1$ regularized linear least squares approximation of the target property (P). The target property is approximated as $P(d) = dc$, where $c$ is the coefficient (or weight) associated with $\Omega$ dimensional descriptor $d$. The solution to this equation can be determined by minimizing the *argmin* $(||P - Dc||_2^2) + \lambda||c||_1$.

The coefficient $c$ is non-zero for all featurized descriptor which is then ranked to determine the corresponding importance.

The generation of nonlinear combinations of descriptors ($d_i$) by applying several mathematical operators such as, *1/x, √x, x², x³, log(x), 1/ log(x),* and *exp(x)*, on each feature allows to form a nonlinear mapping between $D$ and $P$. In addition, the complexity by combining these functionalized descriptors via summation allows to generate a more effective map between $D$ and $P$. We have considered functionalized descriptors utilizing 2 or 3 terms for the purpose of this study. Here, we note that inclusion of more terms may lead to more accurate correlation to endpoint. However, it also introduces additional uncertainty carried by each of the terms combined with mathematical operators. The physics-informed featurization and sparsification method allows us to combine multiple features in a linear combination to establish direct correlation to the endpoint target. Within this method, we search over a large combinatorial space, combine features followed by balancing units/dimensionality (with coefficients) to convert them into feasible equations. These are the mathematical formulations that are then turned into probabilistic models (hypotheses) by introducing suitable priors on parameters, applicable to all use cases.

The second element of the proposed approach is the hypothesis-driven active learning built upon SISSO-derived functional forms. In the hypothesis learning, we utilize a *structured* GP (*s*GP) as opposed to standard zero mean GP as our surrogate model(s) to insert physics-informed priors in the GP/BO framework. To illustrate this approach, we note that in the conventional GP/BO process, GP is defined as

$$y = f(x) + \varepsilon,$$
$$f \sim MVNormal\,(m, K)$$

(1)

where *MVNormal* is a multivariate normal distribution, $m$ is a prior mean function typically set to 0, $K$ is a prior covariance functions (kernel), and $\varepsilon$ is a normally distributed observational noise.

The training process of GP model involves inferring kernel parameters given the available set of observations $(x, y)$ using Bayesian inference techniques. Once the training is completed, the probabilistic predictions of the function over the unmeasured parameter space can be obtained by sampling from a distribution:

$$f_* \sim MVNormal\left(\mu_\theta^{post}, \Sigma_\theta^{post}\right)$$

(2)

$$\mu_\theta^{post} = m(X_*) + K(X_*, X|\theta)K(X, X|\theta)^{-1}(y - m(X)),$$
$$\Sigma_\theta^{post} = K(X_*, X_*|\theta) - K(X_*, X|\theta)(X, X|\theta)^{-1}K(X, X_*|\theta)$$

(3)

Here new inputs are denoted by $X_*$. We can obtain a posterior predictive distribution for each set of kernel parameters. The next point to evaluate is then determined by

$$x_{next} = \arg max_x \frac{1}{M}\sum_{i=1}^{M} \alpha\left(\mu_{\theta^i}^{post}, \Sigma_{\theta^i}^{post}\right)$$

(4)

Here $\alpha$ is a pre-defined acquisition function and $M$ is the number of posterior samples with kernel parameters.

Within the *s*GP, the prior mean function in Equation 1 is substituted by a physics-informed probabilistic model whose parameters are inferred jointly with the kernel parameters. The posterior mean function in Equation 3 then becomes

$$\mu_{\theta^i \phi^i}^{post} = m(X_*|\phi^i) + K(X_*, X|\theta^i) K(X, X|\theta^i)^{-1} \left( y - m(X|\phi^i) \right)$$

(5)

Hence, the active learning scheme with *s*GP incorporates the uncertainty associated with the model parameters. The parametric nature of the model captures the physics-based knowledge to determine the next point of exploration.

The hypothesis-driven learning utilizes an ensemble of the *s*GP models that are effectively competing to reconstruct physical behavior over a large parameters space. The prior mean functions are derived from SISSO method. We utilize Matern kernel for all the *s*GP models and use the predictive uncertainty as the acquisition function, $\alpha = \text{diag}(\Sigma_\theta^{post})$ A basic reinforcement learning policy (epsilon-greedy) is used to sample a hypothesis at each exploration step. The sampled hypothesis gets wrapped into *s*GP whose parameters are inferred using Hamiltonian Monte Carlo. The posterior distributions over the *s*GP parameters are used to obtain a probabilistic prediction over the unmeasured parts of the parameter space. The total predictive uncertainty is then compared with that from a previous step, and the sampled model gets rewarded (penalized) if the uncertainty decreased (increased).

**Formation enthalpy of molecules**:

Predicting molecular energetics to understand formability and stability of isolated molecules have become one of the top applications of modern day's ML models. These molecules play key role in both materials design and drug discovery. Quantum mechanical (QM) calculations within various approximations serve as the basis for studying underlying chemistry, finding active sites to understanding chemical reactions. Among various freely accessible online datasets owning such information, QM9 is popular among researchers due to data homogeneity, purity, and lack of noise. As a result, it has turned out to be a classical benchmark for testing various ML models. QM9 dataset contains information on geometric, energetic, electronic, and thermodynamic properties of ~130,000 organic compounds with elements C, H, O, N and F, as computed with molecular DFT with B3LYP/6- 31G (2df, p) level of theory. Common ML algorithms such as linear regression, kernel ridge regression (KRR), random forest regression and deep neural networks[88] have been used to predict atomization energy,[89] HOMO-LUMO energy,[90] formation energy.[91]

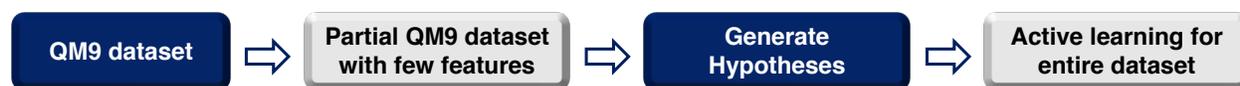

**Figure 2**: **Flow diagram for QM9 dataset**. Figure shows the key stages of the framework as implemented for the QM9 dataset. The key steps include generation of a partial dataset with a combination of a few features from the original QM9 dataset, followed by formulation of scientific hypotheses by physics-informed featurization method which get evaluated in the reward-driven hypothesis-driven learning scheme.

Proper benchmarking investigations,[92-99] comparing computational accuracy, time, convergence of algorithms, starting from traditional ML to state-of-the-art neural networks (NNs) are also available in literature. Most of these studies are dependent on either high dimensional feature space or molecular representations to reach target property. In general, an 80/20 split between the training and test is assumed to perform the benchmarking tasks while drawing conclusions on existing structure-property relations.

Despite the extensive effort, it has not been clearly understood how to establish structure-property relationships on unknown chemical space (even within QM9). In other words, for learning about new molecules and their corresponding properties - which is where capability of ML tools can really be tested - we also need to learn physical or phenomenological laws relating variables with the target of interest on-the-fly in the data-efficient manner.

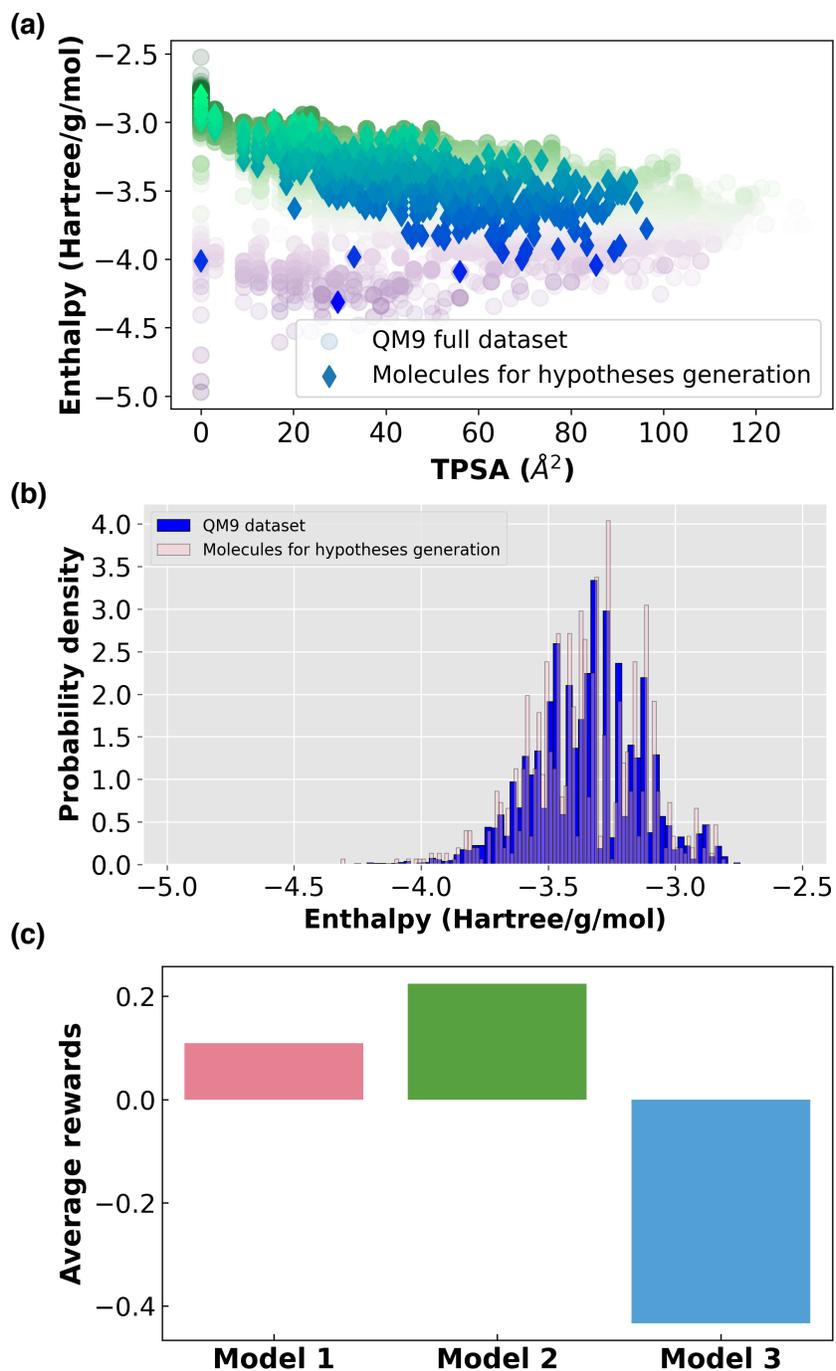

**Figure 3**: **Distribution and prediction of formation enthalpy (Hartree/g/mol)**. (a) Comparative plot and (b) histograms of probability density of formation enthalpy for the entire QM9 dataset superimposed with that of 1000 molecules, used in hypotheses generation. (b) Average rewards for each model predicting formation enthalpy in active learning scheme.

In this example, we lay out how we can employ hypothesis-driven learning to reconstruct behavior of formation enthalpy (FE, Hartree/g/mol) at 298.15 K with respect to topological or polar surface area (TPSA, Å²) for QM9 dataset. A schematic of the framework as implemented for this example is shown in Figure 2. We select the first 1,000 molecules from the QM9 dataset and create a feature-target dataset with a handful of features such as molecular weight (MW), polar surface area (TPSA), molar log P (molelogP), spatial extent (SP) and internal energy (IE). Features such as MW, TPSA, molelogP are computed (easy and computationally cheap to obtain) using the python implementation of RdKit package. The chemical properties such as SP and IE are directly available from QM9 dataset, which are otherwise are calculated using computationally expensive DFT methods.

The general chemistry (as described by the features computed with structural information) of the first 1000 molecules in QM9 largely differ from what is encompassed by the entire dataset. Therefore, the challenge is to find if simple mathematical expressions generated with a small percentage of QM9 are capable to be extended for the remaining data set to yield meaningful predictions of FE. A plot showing distribution of FE of the entire dataset containing 133885 molecules as compared to 0.74% of the dataset as used for generating hypotheses are shown in Figure 3 (a) The normalized histogram plots of FE in Figure 3 (b) show how the first set of 1000 molecules represent the full dataset. We perform the physics-informed featurization to draw three different competing hypotheses as mentioned below.

$$FE = IE * \left(1 + \left(\frac{TPSA}{SP}\right)^2\right)$$
(6)

$$FE = IE * \left(1 + \left(\frac{TPSA}{SP}\right)^2 + molelogP^2\right)$$
(7)

$$FE = IE * \left(1 + \frac{molelogP}{1 + TPSA}\right)$$
(8)

We have considered priors for all parameters as listed below in Table 1. Next, we implement the hypothesis-driven learning scheme by treating the Equations 6-8 as individual probabilistic models wrapped into *s*GP.

**Table 1**: Table lists priors considered for all parameters listed in the equations for predicting excitation energy.

| Parameter | Prior |
| --- | --- |
| IE (Hartree/g/mol) | Uniform [-4, 2] |
| SP (Å²) | Uniform [2, 0.05] |

Molecules utilized at the stage of hypotheses generation are discarded from training and testing. We use 300 samples picked randomly from the dataset as seeds or starting points. Five different set of samples are initialized in this manner. The average reward over all initializations for each model after 200 explorations is shown in Figure 3 (c). Model 2 appears to have accumulated the highest reward to reconstruct FE for the entire chemical space of QM9.

However, if we look at individual runs as plotted in Figure 4, it is evident that Model 1 and Model 2 compete during exploration. The median uncertainty (a, c, e) and reward (b, d, f) for each model with respect to exploration steps for three different random initializations are plotted in Figure 4. The additional jumps in the uncertainty during a segment of exploration steps can be attributed to structural complexity, evolution of the power laws during the learning process. In addition, the approximations given by Model 1 and Model 2 have similar functional forms, due to which, both lead to reasonable predictions. Model 3 always collects the least amount of reward indicating failure of this specific approximation to capture the distribution of FE over the full dataset. The approximations given by the models may slightly vary depending on the initial set of molecules on which the hypotheses are generated. These variations can still be captured by going to higher order approximations during formulation of model expressions which is expected to include combination of the already utilized terms in our present models.

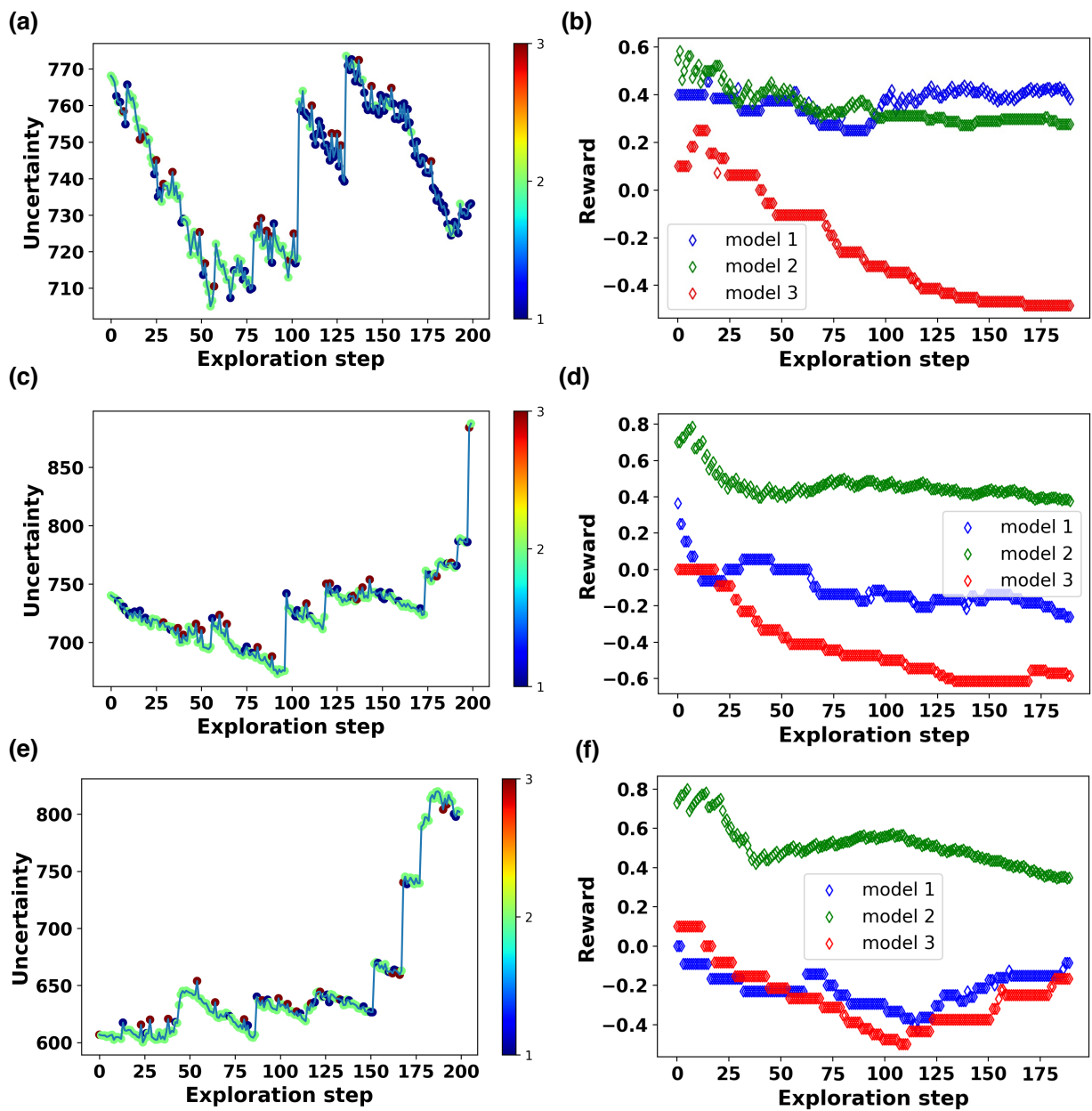

**Figure 4**: **Hypothesis-driven active learning with structured Gaussian processes (sGPs) for entire QM9 dataset**. The median uncertainty (a, c, e) and reward (b, d, f) for each model with respect to explorations step for three different random initializations are shown here.

This example establishes the capability of this technique to reconstruct behavior of a molecular property and how it varies over a wide variety of chemical space. It only utilizes a handful of easily computable features along with simplistic (more interpretable) mathematical formulations, leading to meaningful predictions.

It is important to emphasize that the main goal of building such a framework is to learn or estimate generalized physical laws in an active learning setting. Typically, ML models are assessed based on their performance on test/validation sets, but this approach neglects the impact of real-world data's out-of-distribution characteristics. Consequently, relying on average errors from hold-out validation sets might not provide the most accurate representation of an ML framework's predictive capabilities for unseen data. Instead, focusing on approximating generalized physics laws through observed data enables the framework to better predict the behavior of unexplored chemical spaces.

In summary, we have presented an example for molecules in QM9 dataset to showcase the capability of physics-informed featurization combined with hypothesis-driven active learning for reconstruction of materials property. It helps to understand and approximate the functional behavior of systems belonging to different materials class for which data from simulations or experiments may not be available. The framework proposed in this work allows for a couple of prominent advances in the physical science and ML community. First is its potential to serve as a template to come up with easily interpretable models to represent functionalities, obtained from previous observations. It also attempts to meet both design and discovery requirements, truly needed for facilitating progresses in physical sciences.

**Acknowledgements**

This research is sponsored by the INTERSECT Initiative as part of the Laboratory Directed Research and Development Program of Oak Ridge National Laboratory, managed by UT-Battelle, LLC, for the U.S. Department of Energy under contract DE-AC05-00OR22725. Part of this research was conducted at the Center for Nanophase Materials Sciences, which is a DOE Office of Science User Facility.

**Author contributions**

A.G. proposed, designed the workflow, and wrote the draft of the manuscript. S.V.K. discussed the concept and participated in writing. M.Z. developed the structured GP and hypothesis learning approaches and oversaw the preparation of the manuscript.

**Data Availability**

All datasets as utilized in this work can be accessed via the github repository. https://github.com/aghosh92/SISSO_sGP

**Code Availability**

The workflow can be accessed via the github repository. https://github.com/aghosh92/SISSO_sGP

**Conflict of interest**

The authors declare no competing interests.